\documentclass{ecai} 

\usepackage{latexsym}
\usepackage{amssymb}
\usepackage{amsmath}
\usepackage{amsthm}
\usepackage{booktabs}
\usepackage{enumitem}
\usepackage{graphicx}
\usepackage{color}
\usepackage{natbib}

\usepackage{float}

\usepackage{tabularx}
\usepackage[T1]{fontenc}
\usepackage[utf8]{inputenc}
\usepackage{microtype}
\usepackage{inconsolata}
\usepackage{booktabs}
\newcommand{\smallfootnote}[1]{\footnote{\tiny #1}}
\usepackage{times}
\usepackage{subcaption} 
\usepackage{multirow} 

\newcommand{\BibTeX}{B\kern-.05em{\sc i\kern-.025em b}\kern-.08em\TeX}

\begin{document}


\begin{frontmatter}


\paperid{123} 



\title{Do Large Language Models Understand Morality Across Cultures?}


\author[A]{\fnms{Hadi}~\snm{Mohammadi}\orcid{}\thanks{Corresponding Author. Email: h.mohammadi@uu.nl.}
}
\author[A]{\fnms{Yasmeen}~\snm{F.S.S. Meijer}\orcid{}
}

\author[A]{\fnms{Efthymia}~\snm{Papadopoulou}\orcid{}
}
\author[A]{\fnms{Ayoub}~\snm{Bagheri}\orcid{}} 

\address[A]{Department of Methodology and Statistics, Utrecht University, The Netherlands}


\begin{abstract}
Recent advancements in large language models (LLMs) have established them as powerful tools across numerous domains. However, persistent concerns about embedded biases, such as gender, racial, and cultural biases arising from their training data, raise significant questions about the ethical use and societal consequences of these technologies. This study investigates the extent to which LLMs capture cross-cultural differences and similarities in moral perspectives. Specifically, we examine whether LLM outputs align with patterns observed in international survey data on moral attitudes. To this end, we employ three complementary methods: (1) comparing variances in moral scores produced by models versus those reported in surveys, (2) conducting cluster alignment analyses to assess correspondence between country groupings derived from LLM outputs and survey data, and (3) directly probing models with comparative prompts using systematically chosen token pairs. Our results reveal that current LLMs often fail to reproduce the full spectrum of cross-cultural moral variation, tending to compress differences and exhibit low alignment with empirical survey patterns. These findings highlight a pressing need for more robust approaches to mitigate biases and improve cultural representativeness in LLMs. We conclude by discussing the implications for the responsible development and global deployment of LLMs, emphasizing fairness and ethical alignment.


\end{abstract}

 \end{frontmatter}



\section{Introduction}

Large language models (LLMs) have recently taken center stage in both scientific and public debates due to significant advancements in their performance \citep{Bender:2021}. These models now show great promise for applications ranging from search engines and recommendation systems to automated decision-making tools that deeply influence everyday life. Nonetheless, alongside these impressive capabilities, concerns persist regarding the potential biases LLMs can exhibit, such as gender, racial, and cultural bias.

A primary reason for this risk is that LLMs learn from vast, real-world text datasets that may contain societal and cultural prejudices \citep{karpouzis2024, mishra2024}. Consequently, when large portions of the training data systematically reflect certain groups unfavorably, the resulting language model may replicate or even amplify those biases. Given the growing reliance on LLM-based systems across many fields, this raises important questions about whether these models truly capture the diverse moral perspectives observed in actual human societies.

Despite its importance, the issue of whether LLMs accurately reflect cross-cultural moral judgments has been relatively understudied \citep{Arora:2022, liu2023}. In examining how faithfully LLMs capture moral attitudes that vary across cultural contexts, a key consideration is their ability to replicate both the areas of divergence (where cultures disagree) and similarity (where cultures align) on moral topics. Thus, the central research question is:

\begin{quote}
\emph{To what extent do language models capture cultural diversity and common tendencies regarding topics on which people around the world tend to diverge or agree in their moral judgments?}
\end{quote}

Addressing this question carries both scientific and societal significance. Scientifically, it provides insight into how effectively LLMs, trained primarily on text data, can model complex cultural norms. Societally, ensuring these models reflect actual cross-cultural variation is vital for preventing biased or inaccurate representations of different cultural groups \citep{liu2023}. As LLMs increasingly shape public opinion and decision-making, a mismatch between how cultures truly view moral issues and how models characterize these issues can perpetuate prejudice and unfairness. Conversely, LLMs that accurately capture inter-cultural moral differences and similarities can help reveal common ground and support cross-cultural understanding.

Against this backdrop, the present study focuses on evaluating the extent to which contemporary LLMs mirror the diversity and patterns of moral judgments observed across cultures. Three primary methods are employed:
\begin{enumerate}
    \item \textbf{Comparing Variances}: We compare the variance in model-generated moral judgments with the variance in survey-based moral judgments across countries.
    \item \textbf{Cluster Alignment}: We examine the alignment of model-induced country clusters with empirically derived clusters.
    \item \textbf{Direct Comparative Prompts}: We probe LLMs using tailored prompts to see whether they recognize similarities and differences in moral perspectives between countries.
\end{enumerate}

By using these complementary techniques, this work offers insights into the strengths and limitations of LLMs in depicting cross-cultural moral norms, ultimately informing ongoing discussions about their ethical development and deployment. The remainder of this paper is structured as follows. In Section~\ref{sec:literature-review}, we review related research on cross-cultural moral judgments in LLMs and the issue of bias in these models. Section~\ref{sec:method} describes the data and methods used in our analysis, and Section~\ref{sec:results} details the results. We then discuss key findings in and conclude with final remarks in Section~\ref{sec:conclusion}.

\section{Related work on moral judgment and LLM bias}
\label{sec:literature-review}

\subsection{Cross-Cultural Understanding of Moral Judgments in LLMs}

Moral judgments are evaluations of whether specific actions, intentions, or individuals are morally ``good'' or ``bad,'' and they can vary widely across cultures due to social norms, religious doctrines, and historical influences \citep{Haidt:2001, Shweder:1997}. Broadly speaking, Western, Educated, Industrialized, Rich, and Democratic societies—commonly abbreviated as W.E.I.R.D. in cross-cultural psychology literature \citep{henrich2010weirdest}, tend to prioritize autonomy\footnote{The acronym W.E.I.R.D. is a technical term from cross-cultural psychology used to identify a specific cluster of societies that are overrepresented in psychological research. It was introduced to highlight sampling bias in behavioral sciences and has become standard terminology in the field.}, individual rights, and personal choice, whereas many non-W.E.I.R.D. cultures place a higher emphasis on communal obligations, duty, and spiritual purity \citep{graham2016}. For instance, individuals in W.E.I.R.D. contexts commonly regard sexual behaviors as a matter of personal freedom, while those from more community-oriented cultures may treat the same behaviors as collective moral issues.

Scholars such as \citet{Johnson:22} and \citet{Benkler:2023} refer to this diversity of valid yet conflicting moral values as ``moral value pluralism.'' \citet{kharchenko2024} caution that LLMs often fail to accurately reflect this pluralism, partly because these models are trained on large but not necessarily diverse datasets. \citet{du-2024} likewise note that an overemphasis on English-language training data can overshadow the linguistic and cultural richness of the real world, highlighting the importance of multilingual corpora and larger model sizes. Indeed, \citet{Arora:2022} propose that multilingual LLMs hold promise for capturing cross-cultural values, though the potential lack of diversity within available multilingual data remains a limiting factor.

Consistent with these concerns, \citet{Benkler:2023} argue that most AI systems mirror the dominant values of the culture (often Western) producing the majority of the data. This phenomenon can result in a moral bias, whereby W.E.I.R.D. norms and perspectives are incorrectly treated as universally applicable. Empirical investigations of whether LLMs uphold or correct such biases are limited. Some work suggests that they struggle to reproduce culturally specific moral codes \citep{Arora:2022, Benkler:2023}, while other findings are more optimistic about LLMs' capacity to model cultural diversity \citep{Ramezani:2023, mohammadi2025exploring}. This divergence underscores the importance of continued research on how language models perceive and replicate moral frameworks across various cultures.

\subsection{The Risk of Bias in LLMs}

Bias in LLMs arises when these models inherit or amplify prejudices present in their training datasets. Typically, LLMs learn language representations (or embeddings) by analyzing co-occurrences of words across massive corpora. If these corpora disproportionately depict certain groups or behaviors negatively, the learned representations can perpetuate or exacerbate harmful stereotypes in model outputs \citep{Nemani:2024}.

A well-known example is the gender bias identified in word embeddings, where terms like ``woman'' are closely associated with ``homemaker'' and ``man'' with ``computer programmer'' \citep{bolukbasi2016}. Another instance is GPT-3's tendency to associate ``Muslims'' with violent acts more than ``Christians'' \citep{Johnson:22}. Recent work has shown that while demographic biases influence LLM outputs, content-specific features remain the dominant factor in model predictions \citep{mohammadi2025assessing}. Although ongoing research aims to mitigate bias \citep{mishra2024}, this task remains daunting, as biased outputs can influence everything from public sentiment to automated hiring decisions \citep{Noble:2018}.

For instance, an LLM trained on biased sources might disproportionately recommend men for technical positions, perpetuating gender inequality \citep{bolukbasi2016}. In a similar vein, consistently linking certain religious groups with violence can reinforce negative stereotypes and intensify discrimination. Given these high-stakes consequences, developing models that faithfully capture cultural diversity rather than simplifying or skewing moral perspectives is not merely an academic challenge but a moral and societal imperative \citep{Zou:2018}.

In summary, these two strands of literature, (1) how LLMs handle cross-cultural moral judgments and (2) how bias emerges and persists in LLMs, highlight the need to systematically examine how well these models capture the complexities of moral values across different societies. The following sections detail our data sources and methodological approach to investigating these issues.

\section{Data and methods}
\label{sec:method}

\subsection{Datasets}

The World Values Survey\smallfootnote{\url{https://www.worldvaluessurvey.org/WVSDocumentationWV7.jsp}} (WVS) provides detailed information on people's values across cultures. In this study, we use data from Wave~7 \citep{Haerpfer:2022}, which covers the period 2017--2020. This wave features participants from 55 countries who responded to 19 statements on moral issues (e.g., divorce, euthanasia, political violence, cheating on taxes). The survey was administered in the primary languages of each country, offering multiple response categories.

\begin{figure}[h]
  \centering
  \begin{subfigure}[b]{0.45\columnwidth}
    \centering
    \includegraphics[width=\textwidth]{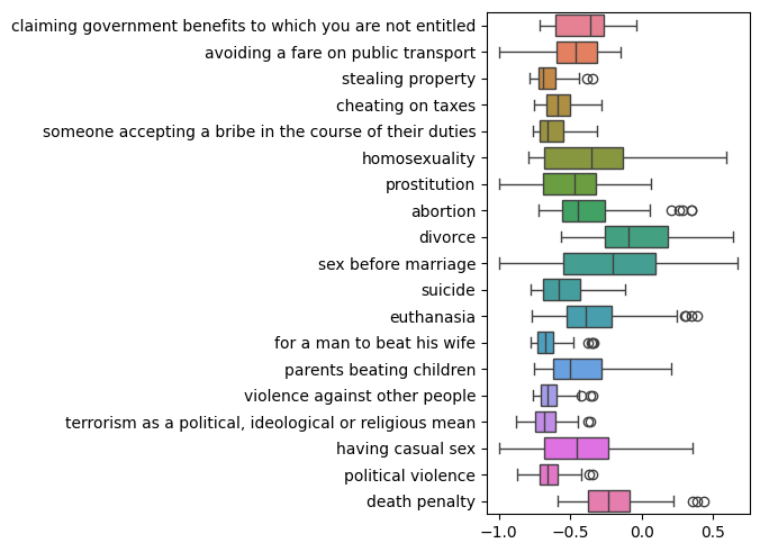}
    \caption{}
    \label{fig:wvs-topics}
  \end{subfigure}
  \hfill
  \begin{subfigure}[b]{0.45\columnwidth}
    \centering
    \includegraphics[width=\textwidth]{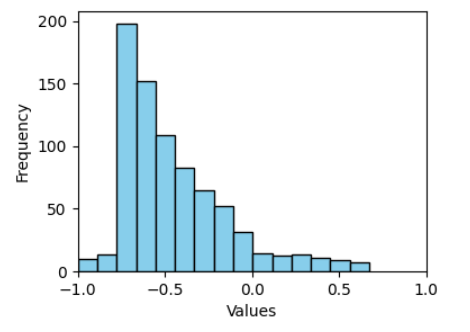}
    \caption{}
    \label{fig:wvs-hist}
  \end{subfigure}
  \vspace{10pt}
  \caption{(a) Spread of responses across moral topics and countries in WVS Wave~7. (b) Distribution of normalized WVS Wave~7 answers.}
  \label{fig:WVS-all}
\end{figure}

Only the country name and each response were retained, with values normalized to range from \([-1,1]\), where \(-1\) indicates “never justifiable” and \(1\) signifies “always justifiable.” These normalized scores facilitate comparability and statistical analysis. For each country–moral issue pair, we computed an average (mean) rating, thus capturing a broad overview of each country's position. We acknowledge that averaging can obscure outlier or minority perspectives, but it was deemed the most feasible approach for this study. 
Figure~\ref{fig:WVS-all} depict the overall distribution of these normalized scores and their variation across topics and countries.

\vspace{10pt}
As a second dataset, we use the Pew Global Attitudes Project\smallfootnote{\url{https://www.pewresearch.org/dataset/spring-2013-survey-data/}} (2013), which surveyed 39 countries (100 participants each) on 8 moral topics, such as drinking alcohol or getting a divorce. The questionnaire was administered in English, allowing respondents to categorize a topic as “morally acceptable,” “not a moral issue,” or “morally unacceptable.”

\begin{figure}[ht]
  \centering
  \begin{subfigure}[b]{0.45\columnwidth}
    \centering
    \includegraphics[width=\textwidth]{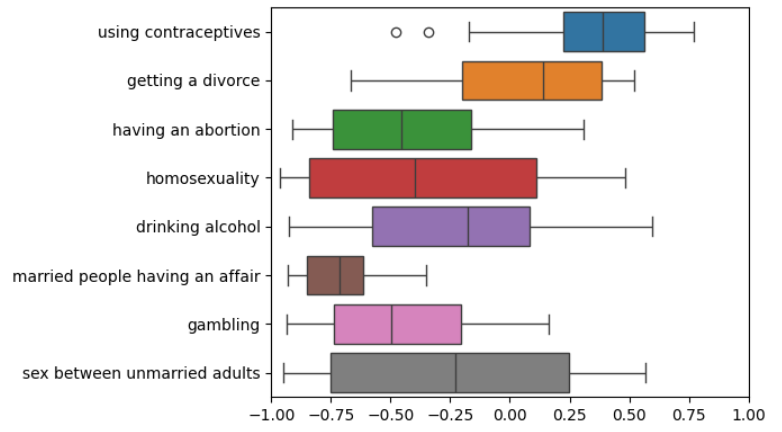}
    \caption{}
    \label{fig:pew-topics}
  \end{subfigure}
  \hfill
  \begin{subfigure}[b]{0.45\columnwidth}
    \centering
    \includegraphics[width=\textwidth]{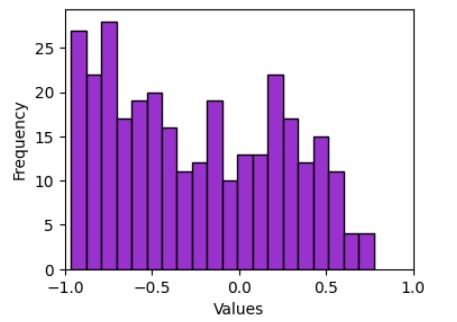}
    \caption{}
    \label{fig:PEW-hist}
  \end{subfigure}
  \vspace{10pt}
  \caption{(a) Spread of responses across moral topics and countries for PEW 2013. (b) Distribution of normalized PEW 2013 responses.}
  \label{fig:PEW-all}
  \vspace{15pt}
\end{figure}

We extracted only country names and responses (Q84A--Q84H), again transforming them to a \([-1,1]\) scale and averaging scores by country–topic pair. Figure~\ref{fig:PEW-all} summarizes the distribution of normalized scores and their topic-level variation.

\subsection{Pre-processing}
In the preprocessing of version 5 of the World Values Survey (WVS) data, the dataset was initially filtered to retain only the columns corresponding to the moral questions Q177 to Q195 and the country code (B\_COUNTRY). These questions cover a range of moral issues, such as tax cheating, accepting bribes, and attitudes towards homosexuality. Following the initial filtering, country names were assigned to each row based on the B\_COUNTRY codes using a predefined country mapping dataset. Responses with values of -1, -2, -4, and -5, which represent 'Don't know,' 'No answer,' 'Not asked in survey,' and 'Missing; Not available,' respectively, were replaced with zero. This adjustment was made to ensure that calculations, such as averaging, were not affected by non-responses. The decision to replace with 0 ensures that the structure of the dataset remains intact. It avoids introducing NaN values or leaving cells empty, which could complicate subsequent data analysis tasks such as averaging or statistical modeling. Moreover, a replacement value of 0 ensures that non-responses do not influence the computed averages or other aggregated measures artificially.  
After replacing non-response values with 0, the dataset was aggregated by country, calculating the mean response for each moral question per country. This provided a country-specific average score for each ethical issue. To enable comparisons across different countries and questions, these average scores were normalized on a scale from -1 to 1, where 1 signifies that the behavior is justifiable in every case and -1 denotes it is never justifiable. This normalization involved adjusting the mean responses, which initially ranged from 1 to 10, to fit the new scale. This step was needed for cross-national comparisons. Finally, normalized values were rounded to four decimal places to enhance clarity.

\subsection{Models}
We begin with two monolingual English models. The first is \textbf{GPT-2}, chosen for its strong performance in generating coherent, contextually relevant text \citep{Radford2019LanguageMA}. We use two versions from Hugging Face,\emph{GPT-2 Medium} (355M parameters) and \emph{GPT-2 Large} (774M parameters), to observe how increased model size influences their capacity to interpret morally charged content. Larger models generally capture more complex patterns and may better approximate cultural moral judgments.  As a second monolingual model, we employ the \textbf{OPT} series by Meta AI \citep{zhang2022opt}. Two variants, \emph{OPT-125M} and \emph{OPT-350M}, are included to benchmark smaller, computationally efficient architectures against larger ones. OPT models, like GPT-2, generate text by predicting the next word in a sequence, having been trained on diverse English corpora. 

Next, we incorporate multilingual models to explore how exposure to varied linguistic data might shape moral judgments across different countries.  The first is \textbf{BLOOM}, a transformer-based, autoregressive language model from the BigScience project, trained on 46 natural and 13 programming languages \citep{Scao2022BLOOMA1}. We specifically use BLOOM-560M and BLOOMZ-560M (fine-tuned for zero-shot learning), rather than the full 176B version, to keep computational requirements manageable. BLOOM’s design aims for strong cross-lingual performance, offering a flexible approach for text tasks in multiple languages. Lastly, we include \textbf{Qwen}, developed by Alibaba Cloud. Qwen is also a multilingual transformer trained on 29 languages (including English and Chinese). Its latest versions demonstrate competitive results in language understanding, multilingual tasks, coding, and reasoning. For feasibility, we use the 0.5B parameter variant in this study.

While hundreds of open-source LLMs currently exist, our model selection was guided by several criteria designed to ensure a representative sample of the current landscape. We prioritized architectural diversity by including both traditional autoregressive models (GPT-2 and OPT) and multilingual transformer models (BLOOM and Qwen), allowing us to test whether different architectural approaches affect cultural understanding. The selected models also span a range of sizes from 125M to 774M parameters, enabling us to investigate whether increased model capacity correlates with better cross-cultural moral representation. Furthermore, we deliberately contrasted monolingual English models with multilingual ones to test the hypothesis that exposure to diverse languages during training might improve cultural sensitivity. All selected models are openly available, ensuring our experiments can be replicated and extended by other researchers. Finally, these specific models have been extensively studied in prior cultural bias research \citep{Arora:2022, Johnson:22}, allowing our findings to build upon and be compared with existing literature. While we acknowledge this represents only a subset of available models, we argue it captures key variations in architecture, size, and training approach that are most relevant to understanding cultural bias. Future work should certainly expand this analysis to newer model families such as LLaMA, Mistral, and other emerging architectures to validate and extend our findings.

\subsection{Evaluation metrics}
We employ three methods to evaluate whether the selected LLMs capture variations and agreements in moral views across cultures: 
\begin{enumerate}
    \item Comparing the variances of survey-based and model-generated moral scores across countries.
    \item Assessing alignment between country clusters derived from these respective scores.
    \item Probing LLMs with direct comparative prompts to evaluate how they recognize cross-cultural moral differences and similarities.
\end{enumerate}

For the first two methods, we compute a model-generated moral score for each country–topic pair by prompting the LLM and examining its log probabilities \citep{mohammadi2025exploring}. Two prompt templates are used:

\begin{itemize}
    \item \texttt{In \{country\} \{topic\} is \{moral\_judgment\}.}
    \item \texttt{People in \{country\} believe \{topic\} is \{moral\_judgment\}.}
\end{itemize}

\begin{table}[h]
\centering
\small
\begin{tabular}{@{}ll@{}}
\toprule
\textbf{Pair \#} & \textbf{Contrasting Statements} \\ 
\midrule
1 & \textit{always justifiable} / \textit{never justifiable} \\
2 & \textit{right} / \textit{wrong} \\
3 & \textit{morally good} / \textit{morally bad} \\
4 & \textit{ethically right} / \textit{ethically wrong} \\
5 & \textit{ethical} / \textit{unethical} \\
\bottomrule
\end{tabular}
\vspace{5pt}
\caption{Token pairs used to prompt the model for moral judgments.}
\label{tab:token-pairs}
\end{table}

As shown in Table~\ref{tab:token-pairs}, five contrasting token pairs (e.g., \emph{always justifiable} vs.\ \emph{never justifiable}) are employed to elicit the LLM’s stance. When probing, for example, the moral score on “abortion” in the United States using the first token pair, we issue:
\begin{quote}
    \small
    \emph{In the United States abortion is always justifiable}\\
    \emph{In the United States abortion is never justifiable}
\end{quote}
The LLM outputs log probabilities for each statement. We subtract the log probability of the “immoral” statement from that of the “moral” statement to obtain a pair-specific score. We do this for all five token pairs in both prompt styles and average the results to produce a final model-generated moral score. This score mirrors the format of the empirical survey-based scores.


\subsection{Evaluation}

\subsubsection{Comparison of Variances}
We first compare cross-cultural variances in moral scores. For each topic, we compute the variance of survey-based scores and model-generated scores, identifying high-variance (controversial) and low-variance (widely agreed upon) topics. We then assess the association between these two sets of variances via Pearson’s correlation. A strong positive correlation suggests the model effectively captures cross-country variability; a weak or negative correlation suggests otherwise \citep{Ramezani:2023}. This step also helps compare which topics surveys and models respectively deem most controversial or consensual.

\subsubsection{Cluster Alignment}
Next, we evaluate whether models replicate empirical cross-country patterns of moral judgment. We cluster countries based on their survey scores using K-means, determining the optimal number of clusters $K$ via repeated runs and selecting the highest silhouette score. The same $K$ is then applied to model-generated scores. We compare these two clusterings using Adjusted Rand Index (ARI) and Adjusted Mutual Information (AMI) \citep{Nazaretsky2020, lazarenko2021}, as well as their average (CAS), to quantify how well the model-based clusters align with the survey-based “ground truth.” 

We repeat this procedure thrice for each dataset: once using all moral topics, once focusing on the most controversial topics, and once on those with highest agreement. This reveals how well the models capture both contentious areas and common ground across cultures.

\subsubsection{Probing with Direct Comparative Prompts}
Finally, we use explicit pairwise comparisons to assess whether models recognize similarities and differences in moral judgments across clusters. For each topic, we apply hierarchical clustering to the survey-based scores, then isolate the two most divergent clusters (those with the largest difference in mean moral score). We repeatedly select representative countries from each cluster and prompt the model using a structure of the form:
\begin{quote}
\small
\emph{Regarding the morality of \{topic\}, the judgments of people in \{country x\} and \{country y\} are \{comparative token\}.}
\end{quote}
where \{comparative token\} takes one of the pairs from Table~\ref{tab:token-pairs} (e.g., \emph{similar}/\emph{dissimilar}). We compare the model’s predicted label, “similar” or “different”, with the empirical classification (same cluster vs.\ different clusters). Confusion matrix metrics and a Chi-squared test of association then measure how closely the model’s judgments align with the survey-based groupings.

\raggedbottom

\section{Results}
\label{sec:results}


To evaluate how well language models capture cross-cultural moral variability, we compare the topic-level variance from two survey datasets (WVS and PEW) with the variance of the corresponding model-generated moral scores. Table~\ref{tab:combined_correlation_variances} summarizes the Pearson correlation (\emph{r}) values, with associated \emph{p}-values, for each model across both datasets. 

\begin{table}[t]
\centering
\renewcommand{\arraystretch}{1}
\footnotesize
\begin{tabular}{lcc|cc}
\toprule
\multirow{2}{*}{\emph{Model}} & \multicolumn{2}{c|}{WVS} & \multicolumn{2}{c}{PEW}\\
\cmidrule(lr){2-3}\cmidrule(lr){4-5}
& \emph{r} & \emph{p} & \emph{r} & \emph{p} \\
\midrule
GPT-2 Medium  & -0.195 & 0.424 & -0.090 & 0.832 \\
GPT-2 Large   & -0.115 & 0.640 &  0.617 & 0.103 \\
OPT-125       & -0.035 & 0.887 & -0.095 & 0.822 \\
QWEN          & -0.200 & 0.413 &  0.102 & 0.811 \\
BLOOM         & -0.118 & 0.631 &  0.608 & 0.110 \\
\bottomrule
\end{tabular}
\vspace{7pt}
\caption{Correlation between topic variances (WVS and PEW) and model-generated moral score variances. None of the correlations reach statistical significance (all \emph{p}~>~0.05).}
\label{tab:combined_correlation_variances}
\end{table}

\paragraph{WVS Variance Correlations.}
The weak negative correlations for all models on the WVS dataset indicate that the model-generated variance does not align with the observed cross-cultural diversity in these topics. Specifically, there is no statistically significant evidence that LLMs capture the degree of controversy reflected in WVS responses. The negative but insignificant correlations highlight how these models fall short in capturing the full range of intercultural nuance.

\paragraph{PEW Variance Correlations.}
On the PEW dataset, correlations are slightly more favorable for GPT-2 Large (\emph{r}~=~0.617) and BLOOM (\emph{r}~=~0.608), suggesting a somewhat better capability to capture topic-level variability. However, even these moderate-to-strong relationships do not achieve statistical significance. In sum, no model consistently reproduces the magnitude of cross-cultural disagreement measured by the PEW data.


Table~\ref{tab:combined_moral_variation_scores} compares the empirical mean moral scores and variance with those generated by each model. We observe a consistent tendency across both WVS and PEW for the models to assign higher mean moral scores (i.e., more \emph{morally acceptable}) and systematically lower variance than in the survey data. This pattern underscores the models’ tendency to view topics as more morally approved and less controversial than they are in reality.

\begin{table}[t]
\renewcommand{\arraystretch}{1.5}
\centering
\scriptsize
\begin{tabular}{lcccc}

\toprule
\multirow{2}{*}{\emph{Source}} & \multicolumn{2}{c}{WVS} & \multicolumn{2}{c}{PEW}\\
\cmidrule(lr){2-3}\cmidrule(lr){4-5}
& \emph{Mean score} & \emph{Var.} & \emph{Mean score} & \emph{Var.} \\
\midrule
\textbf{Empirical} & -0.576 & 0.075 & -0.244 & 0.138 \\
BLOOM              &  0.474 & 0.004 &  0.246 & 0.006 \\
OPT-125            &  0.104 & 0.012 &  0.248 & 0.027 \\
QWEN               &  0.242 & 0.021 &  0.221 & 0.019 \\
GPT-2 Large        &  0.323 & 0.015 &  0.160 & 0.032 \\
GPT-2 Medium       &  0.411 & 0.013 &  0.227 & 0.024 \\
\bottomrule
\end{tabular}
\vspace{7pt}
\caption{Mean moral scores and variances for WVS and PEW topics compared with model-generated values.}
\label{tab:combined_moral_variation_scores}
\end{table}

These lower variances suggest that the models underestimate the degree of cultural disagreement, especially on polarizing issues such as sexuality and family-related norms.

\begin{figure*}
    \centering
    \includegraphics[width=\linewidth]{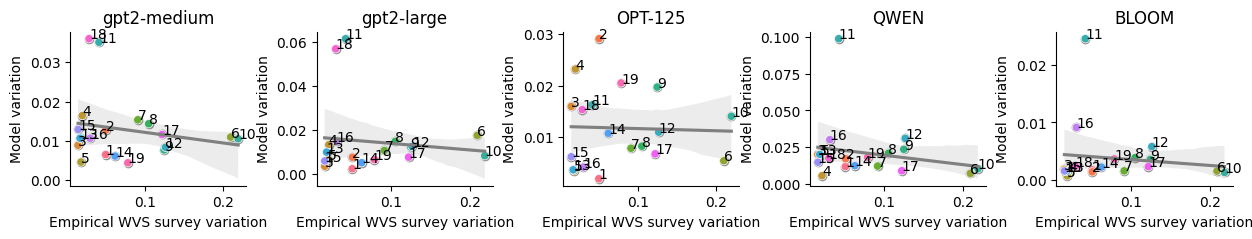}
    \caption{Comparison of empirical and model-inferred moral score variances for WVS topics. The models underestimate cross-cultural disagreements.}
    \label{fig:comparison_wvs}
\end{figure*}
\begin{figure*}
    \centering
    \includegraphics[width=\linewidth]{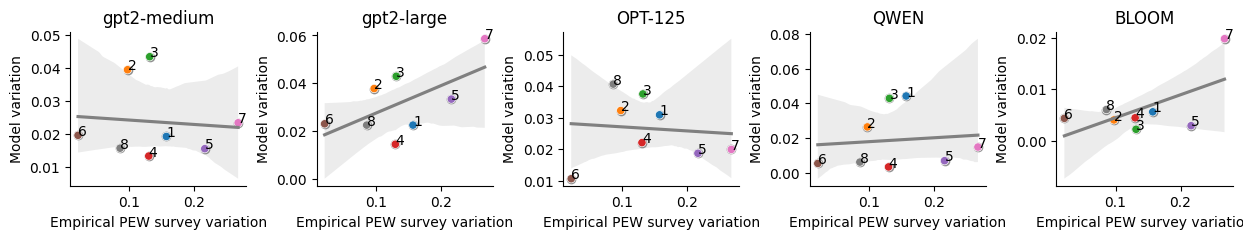}
    \caption{Comparison of empirical and model-inferred moral score variances for PEW topics. Again, models generally exhibit lower variance.}
    \label{fig:comparison_pew}
\vspace{10pt}
\end{figure*}

Figures~\ref{fig:comparison_wvs} and \ref{fig:comparison_pew} illustrate the mismatch between empirical and model-inferred moral variance. Although full rankings for each model are provided in the appendix, Table~\ref{tab:controversy_agreement} summarizes the top three most controversial and most agreed-upon topics, based on \emph{empirical} data from WVS and PEW.

\begin{table}[t]
\vspace{2pt}
\centering
\small
\resizebox{\columnwidth}{!}{%
\begin{tabular}{ll|ll}
\toprule
\multicolumn{2}{c|}{\textbf{WVS}} & \multicolumn{2}{c}{\textbf{PEW}} \\
\midrule
\emph{Topic} & \emph{Var.} & \emph{Topic} & \emph{Var.}\\
\midrule
\textbf{Most controversial} &&& \\
Sex before marriage & 0.219 & Sex between unmarried adults & 0.268 \\
Homosexuality       & 0.209 & Homosexuality               & 0.216 \\
Euthanasia          & 0.126 & Drinking alcohol            & 0.157 \\
\midrule
\textbf{Most agreed upon} &&& \\
Stealing property             & 0.015 & Married people having an affair & 0.021 \\
Violence against other people & 0.015 & Using contraceptives           & 0.086 \\
For a man to beat his wife    & 0.018 & Gambling                        & 0.097 \\
\bottomrule
\end{tabular}}
\vspace{7pt}
\caption{Top three most controversial and most agreed-upon topics from WVS (left) and PEW (right) empirical data.}
\label{tab:controversy_agreement}
\end{table}


From Table~\ref{tab:controversy_agreement}, \emph{sex before marriage} and \emph{homosexuality} rank among the most polarizing topics in both datasets, with variances of 0.219 and 0.209 respectively in WVS, and 0.268 and 0.216 in PEW. These high variances indicate substantial cross-cultural disagreement on these topics, which aligns with prior literature suggesting that sexual and family-related moral issues often reflect deep cultural differences between societies that prioritize individual autonomy versus those emphasizing communal values and traditional norms \citep{graham2016, Shweder:1997}. The fact that these particular topics show the highest variance suggests they serve as key differentiators between moral frameworks across cultures. However, several models misjudge at least one of these issues as relatively uncontroversial, with QWEN and BLOOM even ranking homosexuality among their most agreed-upon topics (as shown in the appendix), suggesting they fail to capture these fundamental cultural divisions.

Although GPT-2 Large and BLOOM show moderate correlations in the PEW dataset (Table~\ref{tab:combined_correlation_variances}), no model achieves statistically significant alignment with the empirical data. Across both WVS and PEW, language models:
\begin{enumerate}[noitemsep,leftmargin=1.5em]
    \item Overestimate moral acceptability, assigning more positive moral judgments to most topics.
    \item Underestimate the degree of cultural disagreement, producing lower variance scores.
\end{enumerate}

These findings suggest that current LLMs do not yet mirror real-world moral heterogeneity, especially for hotly debated topics like sexual and family norms. Simply increasing model size may be insufficient; more nuanced training or alignment with culturally diverse data sources may be necessary to capture the complexity seen in empirical moral attitudes.

\begin{table*}[t]
\centering
\footnotesize
\setlength{\tabcolsep}{4pt}
\begin{tabularx}{\linewidth}{l l r r r r l r}
\toprule
\textbf{Model} & \textbf{Survey} & \textbf{Survey Var.} & \textbf{Survey Mean} & \textbf{Model Var.} & \textbf{Model Mean} & \textbf{Topic} & \textbf{Var. Diff} \\
\midrule
GPT-2 Medium
  & WVS & 0.219 & -0.244 & 0.011 & 0.465 & sex before marriage               & 0.208 \\
  & WVS & 0.209 & -0.396 & 0.011 & 0.577 & homosexuality                     & 0.198 \\
  & WVS & 0.126 & -0.430 & 0.008 & 0.481 & euthanasia                        & 0.118 \\
  & WVS & 0.125 & -0.150 & 0.008 & 0.217 & divorce                           & 0.117 \\
  & WVS & 0.122 & -0.452 & 0.012 & 0.371 & having casual sex                 & 0.110 \\
  & PEW & 0.268 & -0.219 & 0.023 & 0.044 & sex between unmarried adults      & 0.244 \\
  & PEW & 0.216 & -0.342 & 0.016 & 0.641 & homosexuality                     & 0.201 \\
  & PEW & 0.157 & -0.234 & 0.019 & 0.142 & drinking alcohol                  & 0.138 \\
\midrule
GPT-2 Large
  & WVS & 0.219 & -0.244 & 0.008 & 0.454 & sex before marriage               & 0.211 \\
  & WVS & 0.209 & -0.396 & 0.018 & -0.086 & homosexuality                    & 0.192 \\
  & WVS & 0.122 & -0.452 & 0.008 & 0.470 & having casual sex                 & 0.114 \\
  & WVS & 0.126 & -0.430 & 0.013 & 0.261 & euthanasia                        & 0.114 \\
  & WVS & 0.125 & -0.150 & 0.012 & 0.121 & divorce                           & 0.112 \\
  & PEW & 0.268 & -0.219 & 0.059 & -0.138 & sex between unmarried adults     & 0.209 \\
  & PEW & 0.216 & -0.342 & 0.033 & -0.188 & homosexuality                    & 0.183 \\
  & PEW & 0.157 & -0.234 & 0.023 & 0.210 & drinking alcohol                  & 0.135 \\
\midrule
OPT-125
  & WVS & 0.219 & -0.244 & 0.014 & 0.475 & sex before marriage               & 0.205 \\
  & WVS & 0.209 & -0.396 & 0.005 & 0.255 & homosexuality                     & 0.204 \\
  & WVS & 0.126 & -0.430 & 0.011 & 0.013 & euthanasia                        & 0.115 \\
  & WVS & 0.122 & -0.452 & 0.007 & 0.093 & having casual sex                 & 0.115 \\
  & WVS & 0.125 & -0.150 & 0.020 & -0.261 & divorce                          & 0.105 \\
  & PEW & 0.268 & -0.219 & 0.020 & 0.512 & sex between unmarried adults      & 0.248 \\
  & PEW & 0.216 & -0.342 & 0.019 & 0.570 & homosexuality                     & 0.198 \\
  & PEW & 0.157 & -0.234 & 0.031 & 0.187 & drinking alcohol                  & 0.126 \\
\midrule
QWEN
  & WVS & 0.219 & -0.244 & 0.010 & 0.415 & sex before marriage               & 0.209 \\
  & WVS & 0.209 & -0.396 & 0.007 & 0.466 & homosexuality                     & 0.202 \\
  & WVS & 0.122 & -0.452 & 0.009 & 0.177 & having casual sex                 & 0.113 \\
  & WVS & 0.125 & -0.150 & 0.024 & -0.042 & divorce                          & 0.101 \\
  & WVS & 0.126 & -0.430 & 0.031 & -0.115 & euthanasia                       & 0.095 \\
  & PEW & 0.268 & -0.219 & 0.015 & 0.494 & sex between unmarried adults      & 0.253 \\
  & PEW & 0.216 & -0.342 & 0.007 & 0.562 & homosexuality                     & 0.209 \\
  & PEW & 0.130 & -0.405 & 0.004 & 0.130 & having an abortion                & 0.127 \\
\midrule
BLOOM
  & WVS & 0.219 & -0.244 & 0.001 & 0.662 & sex before marriage               & 0.218 \\
  & WVS & 0.209 & -0.396 & 0.002 & 0.865 & homosexuality                     & 0.208 \\
  & WVS & 0.124 & -0.150 & 0.004 & 0.569 & divorce                           & 0.121 \\
  & WVS & 0.126 & -0.429 & 0.006 & 0.712 & euthanasia                        & 0.121 \\
  & WVS & 0.122 & -0.452 & 0.002 & 0.422 & having casual sex                 & 0.120 \\
  & PEW & 0.268 & -0.219 & 0.020 & 0.374 & sex between unmarried adults      & 0.248 \\
  & PEW & 0.216 & -0.342 & 0.003 & 0.843 & homosexuality                     & 0.213 \\
  & PEW & 0.157 & -0.234 & 0.006 & 0.159 & drinking alcohol                  & 0.152 \\
\bottomrule
\end{tabularx}
\vspace{5pt}
\caption{Variance gaps between survey data and model outputs (WVS vs. PEW), showing the top eight topic–model pairs with the largest differences. Full results in the Appendix.}
\label{tab:variance_gaps}
\end{table*}

\subsection{Cluster Alignment}
\label{sec:cluster_alignment}

We analyze how closely the clusters induced by model-generated moral scores align with the empirical clusters derived from both the WVS and PEW datasets. Three metrics are used to measure this alignment: the Adjusted Rand Index (ARI), the Adjusted Mutual Information (AMI), and the Combined Alignment Score (CAS). Higher values on these metrics suggest better agreement between the empirical clusters and the model-generated clusters.


Table~\ref{tab:cluster_alltopics_combined} combines the alignment scores for all topics in WVS (left panel) and PEW (right panel). QWEN shows notably higher metrics on WVS than the other models, indicating closer alignment to empirical scores. For PEW, GPT-2 Large and OPT-125 share moderate alignment scores, while QWEN and BLOOM perform relatively worse.

\begin{table}[t]
\vspace{5pt}
\centering
\small
\resizebox{\columnwidth}{!}{%
\begin{tabular}{l|ccc|ccc}
\toprule
\multirow{2}{*}{\textbf{Model}} & \multicolumn{3}{c|}{\textbf{WVS}} & \multicolumn{3}{c}{\textbf{PEW}} \\
\cmidrule(lr){2-4}\cmidrule(lr){5-7}
 & \textbf{ARI} & \textbf{AMI} & \textbf{CAS} & \textbf{ARI} & \textbf{AMI} & \textbf{CAS} \\
\midrule
GPT-2 Medium & -0.012 & -0.002 & -0.007 & 0.087 & 0.068 & 0.078 \\
GPT-2 Large  &  0.028 &  0.040 &  0.034 & 0.129 & 0.123 & 0.126 \\
OPT-125      & -0.073 &  0.037 & -0.018 & 0.129 & 0.123 & 0.126 \\
QWEN         &  0.291 &  0.138 &  0.215 & -0.019 & 0.065 & 0.023 \\
BLOOM        &  0.015 & -0.011 &  0.002 & 0.008  & -0.004 & 0.002 \\
\bottomrule
\end{tabular}}
\vspace{3pt}
\caption{Cluster alignment scores for all topics in WVS (left) and PEW (right).}
\label{tab:cluster_alltopics_combined}
\end{table}


Table~\ref{tab:cluster_controversial_combined} shows the alignment results for the most controversial topics in both WVS and PEW. All models yield negative or near-zero alignment on WVS. On PEW, GPT-2 Medium remains negative while GPT-2 Large, OPT-125, and BLOOM achieve positive scores, with OPT-125 notably highest.

\begin{table}[t]
\vspace{5pt}
\centering
\small
\resizebox{\columnwidth}{!}{%
\begin{tabular}{l|ccc|ccc}
\toprule
\multirow{2}{*}{\textbf{Model}} & \multicolumn{3}{c|}{\textbf{WVS}} & \multicolumn{3}{c}{\textbf{PEW}} \\
\cmidrule(lr){2-4}\cmidrule(lr){5-7}
 & \textbf{ARI} & \textbf{AMI} & \textbf{CAS} & \textbf{ARI} & \textbf{AMI} & \textbf{CAS} \\
\midrule
GPT-2 Medium & -0.015 & -0.011 & -0.013 & -0.026 & -0.019 & -0.022 \\
GPT-2 Large  & -0.012 &  0.023 &  0.005 &  0.093 &  0.081 &  0.087 \\
OPT-125      & -0.021 &  0.017 & -0.002 &  0.131 &  0.140 &  0.136 \\
QWEN         & -0.014 & -0.018 & -0.016 & -0.006 &  0.073 &  0.033 \\
BLOOM        & -0.015 & -0.011 & -0.013 &  0.009 &  0.006 &  0.007 \\
\bottomrule
\end{tabular}}
\vspace{3pt}
\caption{Cluster alignment scores for most controversial topics in WVS (left) and PEW (right).}
\label{tab:cluster_controversial_combined}
\end{table}


Table~\ref{tab:cluster_agreed_combined} reports the alignment results for the most agreed-upon topics. For WVS, GPT-2 Medium and OPT-125 have positive scores, whereas GPT-2 Large, QWEN, and BLOOM remain negative. For PEW, GPT-2 Medium, GPT-2 Large, and OPT-125 show moderate positive alignment; QWEN and BLOOM exhibit minimal or negative scores.

\begin{table}[t]
\centering
\small
\resizebox{\columnwidth}{!}{%
\begin{tabular}{l|ccc|ccc}
\toprule
\multirow{2}{*}{\textbf{Model}} & \multicolumn{3}{c|}{\textbf{WVS}} & \multicolumn{3}{c}{\textbf{PEW}} \\
\cmidrule(lr){2-4}\cmidrule(lr){5-7}
 & \textbf{ARI} & \textbf{AMI} & \textbf{CAS} & \textbf{ARI} & \textbf{AMI} & \textbf{CAS} \\
\midrule
GPT-2 Medium &  0.079 &  0.010 &  0.044 &  0.057 &  0.045 &  0.051 \\
GPT-2 Large  & -0.019 & -0.014 & -0.016 &  0.028 &  0.020 &  0.024 \\
OPT-125      &  0.120 &  0.038 &  0.079 &  0.035 &  0.051 &  0.043 \\
QWEN         & -0.005 & -0.017 & -0.011 & -0.020 & -0.016 & -0.018 \\
BLOOM        & -0.030 & -0.012 & -0.021 &  0.006 &  0.004 &  0.005 \\
\bottomrule
\end{tabular}}
\vspace{5pt}
\caption{Cluster alignment scores for most agreed-upon topics in WVS (left) and PEW (right).}
\vspace{5pt}
\label{tab:cluster_agreed_combined}
\end{table}

\subsection{Probing with Direct Comparative Prompts}
\label{sec:direct_prompts}

We further examine how models recognize similarities and differences in moral judgments by prompting them to compare topics directly. Tables~\ref{tab:combined_confusion} and \ref{tab:combined_chi} show, respectively, the confusion-matrix scores and chi-squared results for WVS (left) and PEW (right).

\begin{table}[t]
\centering
\renewcommand{\arraystretch}{1.2}
\small
\resizebox{\columnwidth}{!}{%
\begin{tabular}{l|cccc|cccc}
\toprule
\multirow{2}{*}{\textbf{Model}} & \multicolumn{4}{c|}{\textbf{WVS}} & \multicolumn{4}{c}{\textbf{PEW}} \\
\cmidrule(lr){2-5}\cmidrule(lr){6-9}
 & \textbf{Acc.} & \textbf{Prec.} & \textbf{Rec.} & \textbf{F1} 
 & \textbf{Acc.} & \textbf{Prec.} & \textbf{Rec.} & \textbf{F1}\\
\midrule
GPT-2 Medium & 0.485 & 0.488 & 0.336 & 0.398 & 0.495 & 0.494 & 0.402 & 0.444 \\
GPT-2 Large  & 0.509 & 0.508 & 0.946 & 0.661 & 0.495 & 0.497 & 0.954 & 0.654 \\
OPT-125      & 0.502 & 0.510 & 0.461 & 0.484 & 0.506 & 0.506 & 0.480 & 0.493 \\
QWEN         & 0.500 & 0.504 & 0.831 & 0.628 & 0.493 & 0.495 & 0.694 & 0.578 \\
BLOOM        & 0.495 & 0.543 & 0.026 & 0.050 & 0.497 & 0.326 & 0.006 & 0.011 \\
\bottomrule
\end{tabular}}
\vspace{5pt}
\caption{Confusion matrix scores from direct probing on WVS (left) and PEW (right). Acc. = accuracy, Prec. = precision, Rec. = recall.}
\label{tab:combined_confusion}
\vspace{5pt}
\end{table}

Accuracy for all models hovers near 0.5. GPT-2 Large and QWEN stand out with high recall (0.946 and 0.831, respectively), but their precision is lower, yielding moderate F1 scores. BLOOM displays poor performance across most metrics, indicating difficulties in classifying positive and negative instances.

Again, overall accuracy remains near 0.5 for all models. GPT-2 Large shows the highest recall (0.954), while QWEN achieves a recall of 0.694. BLOOM exhibits very low recall and precision, resulting in the lowest F1.

\begin{table}[t]
\centering
\small
\resizebox{0.85\columnwidth}{!}{%
\begin{tabular}{l|cc|cc}
\toprule
\multirow{2}{*}{\textbf{Model}} & \multicolumn{2}{c|}{\textbf{WVS}} & \multicolumn{2}{c}{\textbf{PEW}} \\
\cmidrule(lr){2-3}\cmidrule(lr){4-5}
 & \(\chi^2\) & \textbf{p} & \(\chi^2\) & \textbf{p}\\
\midrule
GPT-2 Medium & 8.38  & 0.004** & 0.418 & 0.518 \\
GPT-2 Large  & 1.491 & 0.222   & 3.325 & 0.068 \\
OPT-125      & 0.338 & 0.561   & 0.609 & 0.435 \\
QWEN         & 1.416 & 0.234   & 1.017 & 0.313 \\
BLOOM        & 1.279 & 0.258   & 4.599 & 0.032* \\
\bottomrule
\end{tabular}}
\vspace{5pt}
\caption{Chi-squared test results from direct probing on WVS (left) and PEW (right). (\(**\)) indicates \(p<0.01\), (\(*\)) indicates \(p<0.05\).}
\vspace{5pt}
\label{tab:combined_chi}
\end{table}

Table~\ref{tab:combined_chi} shows that GPT-2 Medium exhibits a significant (\(p<0.01\)) alignment with WVS, implying its judgments correlate with actual moral (dis)similarities. The other models do not significantly align with WVS. For PEW, BLOOM yields a statistically significant (\(p<0.05\)) result—though this may reflect a consistent but incorrect pattern, given its poor F1 and recall.

Although some models (e.g., GPT-2 Large, QWEN) display high recall indirect probing, their precision is often lacking. GPT-2 Medium is uniquely significant in the WVS chi-squared test, while BLOOM is significant in the PEW test but shows low classification performance overall. These divergences suggest that while models capture certain aspects of moral similarity, they struggle to reflect the full complexity of real-world intercultural judgments.

\section{Discussion and conclusion}
\label{sec:conclusion}

The findings of this study shed light on the capability of LLMs to accurately capture cultural diversity and common tendencies across different moral topics. The investigation utilized multiple methodologies that were based on probing LLMs with prompts derived from the World Values Survey (WVS) and PEW datasets, focusing on a range of moral topics.

\subsection{Comparison of variance}
The correlation analysis between model-generated moral scores and empirical survey data revealed mixed results. For the PEW dataset, GPT-2 Large and BLOOM demonstrated moderate to strong alignment in capturing cultural variations. The fact that the largest model (GPT-2 Large) and the largest multilingual model (BLOOM) performed best may suggest that model size and multilinguality have a positive effect on models' ability to grasp patterns of cultural diversity, which would be in line with previous work from \citet{du-2024} and \citet{Arora:2022}. However, the correlations did not reach statistical significance and therefore no strong claims can be made. Moreover, model performance shows high variability, with weak negative correlations observed for both GPT-2 Large and BLOOM when comparing their variances with the WVS moral score variances. The other models performed weakly and variably in both the PEW and WVS moral score variance comparisons. Furthermore, the models struggled to accurately identify the most controversial and agreed on topics. In fact, some of the models incorrectly categorized (one of) the two most controversial topics as among the most agreed on. The variable and low overall performance could be attributed to the fact that the complexity and nuance of moral values across different cultural contexts may not be fully captured by the models' training data.

\subsection{Cluster alignment}
The clustering alignment results further emphasized the variability in model performance. Overall, GPT-2 Large and OPT-125 showed better alignment with empirical moral scores from both datasets quite consistently, suggesting their relative proficiency in clustering countries based on moral attitudes. However, other models, most notably BLOOM, exhibited lower alignment scores, indicating shortcomings in their ability to mirror the clustering patterns observed in the survey data. These results suggest that the models fall short in grasping cultural patterns regarding moral judgments, which is in line with the findings from the previous method. Thus, while GPT-2 Large and OPT-125 generally show better alignment with empirical moral scores across various topics, the variability in model performance underscores the challenges in accurately capturing the complexities of moral attitudes across different cultural contexts. Overall, the clusterings based on the model scores do not faithfully capture the cultural patterns observed in the clusterings derived from the survey scores.

\subsection{Probing with direct comparative prompts} 
Direct probing with comparative prompts provided additional insights into the models' understanding of moral differences between culturally distinct groups. In general, performance is low as the scores are no higher or even slightly lower than random chance. GPT-2 Large and QWEN stood out with higher accuracy and recall scores, indicating their better performance in distinguishing moral differences between the most divergent clusters identified by the survey data. Upon further inspection, however, it became clear that GPT-2 Large and QWEN almost always predict the same class, which does not signify a proper understanding of inter-cultural differences and similarities. If we disregard the performance of GPT-2 Large and QWEN due to the fact that they always predict the same class, GPT-2 Medium and OPT-125 exhibit the most balanced performance across the remaining models. BLOOM exhibited the lowest performance metric scores, suggesting challenges in discerning nuanced moral judgments across cultures. Notably, despite its low overall performance, BLOOM's judgments were found to be statistically associated with the judgments based on the PEW dataset through a Chi-squared test. This suggests that there may be some alignment between BLOOM's outputs and the moral judgments reflected in the PEW dataset. However, it is important to note that this statistical association does not necessarily imply a meaningful understanding or accurate representation of moral differences between cultures.   

\subsection{Conclusion}   
In conclusion, the study underscores the importance of rigorous evaluation methodologies when assessing LLMs' ability to understand and reflect cultural diversity in moral judgments. The tested models seem to propagate a homogenized view on cross-cultural moral values, identifying most topics as cross-culturally agreed on as more morally acceptable than empirically observed. Thereby, the models generally seem to reflect a rather liberal view, in line with the autonomy-endorsing values found in W.E.I.R.D. societies \citep{graham2016}. It has been established in the literature that exclusively English training data plays a big part in the embedding of homogenous W.E.I.R.D. values and, thereby, cultural bias in LLMs \citep{Benkler:2023}. This could lead one to believe that multilingual LLMs are the answer to mitigating bias in LLMs \citep{Arora:2022}. However, this study could not find convincing evidence to suggest that multilingual models are better at truthfully capturing cultural diversities in moral judgments than monolingual models. Similarly, while model size could be considered another factor influencing model performance due to its potential to enhance computational capacity and capture more complex patterns \citep{du-2024}, its impact was not found to be convincing in the carried out analyses. It can be concluded that this study found no remarkable differences between the tested models in their success, regardless of multilinguality or model size. Overall, the models examined show variable performance and generally exhibit low success in aligning with empirical moral data from global surveys. Thus, ongoing research and development are needed to enhance their accuracy and reliability in diverse cultural settings. Addressing these challenges is crucial for ensuring the ethical integrity and societal impact of AI technologies in the context of global applications. 
\subsection{Conclusion}   
In conclusion, the study underscores the importance of rigorous evaluation methodologies when assessing LLMs' ability to understand and reflect cultural diversity in moral judgments. The tested models seem to propagate a homogenized view on cross-cultural moral values, identifying most topics as cross-culturally agreed on as more morally acceptable than empirically observed. Thereby, the models generally seem to reflect a rather liberal view, in line with the autonomy-endorsing values found in Western, Educated, Industrialized, Rich, and Democratic (W.E.I.R.D.) societies \citep{graham2016}. It has been established in the literature that exclusively English training data plays a big part in the embedding of homogenous W.E.I.R.D. values and, thereby, cultural bias in LLMs \citep{Benkler:2023}. This could lead one to believe that multilingual LLMs are the answer to mitigating bias in LLMs \citep{Arora:2022}. However, this study could not find convincing evidence to suggest that multilingual models are better at truthfully capturing cultural diversities in moral judgments than monolingual models.

Based on our findings, several actionable strategies could improve cultural representativeness in LLMs. First, diversifying training data by prioritizing text from underrepresented regions and languages would help counteract the current bias toward W.E.I.R.D. perspectives. This includes incorporating religious texts, local news sources, and cultural forums that discuss moral topics from non-W.E.I.R.D. societies. Second, culture-aware fine-tuning approaches could be developed using datasets that explicitly represent diverse moral perspectives on controversial topics, weighted to reflect actual global population distributions rather than internet data availability. Third, prompt engineering strategies that explicitly invoke cultural context could elicit more culturally diverse responses. For example, prompts like ``From the perspective of someone in [country] with traditional values...'' may help models access different moral frameworks. Finally, establishing standardized evaluation frameworks using surveys like WVS and PEW would enable regular assessment of cultural bias in new models before deployment. These recommendations provide concrete pathways for researchers and practitioners working toward more culturally inclusive AI systems. The challenges identified here align with broader issues in developing transparent and interpretable NLP systems across various domains \citep{mohammadi2025explainability}, emphasizing the need for continued research in explainable AI methods.

\section{Limitations}

While this study provides important insights, it is important to recognize certain boundaries of our approach. First, although WVS and PEW are well-established surveys covering 55 and 39 countries respectively, they organize complex moral views into fixed categories, which may not capture every nuance or implicit aspect of moral reasoning. Additionally, our analysis examined aggregate patterns across all countries rather than country-specific contributions to variance. Future work could benefit from analyzing which specific countries or regions show the largest discrepancies between model outputs and survey responses, which would provide more granular insights into geographical patterns of model bias. Second, we focused on a selected group of models, so our findings primarily reflect these particular architectures. Third, the choice of prompts in our experiments can influence model responses \citep{Wang2024}, meaning that exploring alternative prompt strategies could yield additional insights. Lastly, due to computational limits, we randomly selected topics in Method~3, which may not cover all diversity within each cluster. Future research can build on our work by testing a wider range of models, experimenting with different prompt designs, analyzing country-specific patterns, and using broader topic sampling to further enrich the analysis.

\section{Ethics Statement}
\label{sec:ethics}

The work relies exclusively on two publicly available, anonymised survey 
datasets (WVS Wave 7 and Pew 2013) and on open-access language models. No 
personal or sensitive information was collected, and all analyses were performed 
in accordance with the data providers’ terms of use. By quantifying cultural 
bias in LLMs we aim to support fairer deployment of generative AI and to 
encourage the creation of more globally representative training data.









\begin{ack}
We appreciate the maintainers of WVS and PEW data for enabling large-scale cross-cultural analysis.
\end{ack}



\bibliography{mybibfile}

\appendix
\section{Appendix}

\subsection{\small Most controversial WVS topics according to models}
\vspace{-5pt}
\begin{table}[H]
    \centering
    \small
    \footnotesize
    \begin{tabular}{p{3.75cm}p{2cm}l}
        \toprule
        \emph{Topic} & \emph{Model} & \emph{Variance} \\
        \midrule
        Political violence & GPT-2 Medium & 0.036 \\
        Suicide & GPT-2 Medium & 0.035 \\
        Cheating on taxes & GPT-2 Medium & 0.016 \\
        \bottomrule
    \end{tabular}
    \vspace{2pt}
    \caption{\scriptsize Top 3 most controversial WVS topics according to GPT-2 Medium}
    \label{tab:controversial_topicsWVSgpt2-m}
\end{table}

\begin{table}[H]
    \centering
    \small
    \footnotesize
    \begin{tabular}{p{3.75cm}p{2cm}l}
        \toprule
        \emph{Topic} & \emph{Model} & \emph{Variance} \\
        \midrule
        Suicide & GPT-2 Large & 0.062 \\
        Political violence & GPT-2 Large & 0.057 \\
        Homosexuality & GPT-2 Large & 0.018 \\
        \bottomrule
    \end{tabular}
    \vspace{2pt}
    \caption{\scriptsize Top 3 most controversial WVS topics according to GPT-2 Large}
    \label{tab:controversial_topics_WVSgpt2-l}
\end{table}

\begin{table}[H]
    \centering
    \footnotesize
    \begin{tabular}{p{4cm}p{2cm}l}
        \toprule
        \emph{Topic} & \emph{Model} & \emph{Variance} \\
        \midrule
        Avoiding a fare on public transport & OPT-125 & 0.029 \\
        Cheating on taxes & OPT-125 & 0.023 \\
        Death penalty & OPT-125 & 0.021 \\
        \bottomrule
    \end{tabular}
    \vspace{2pt}
    \caption{\scriptsize Top 3 most controversial WVS topics according to OPT-125}
    \label{tab:controversial_topicsWVSopt125}
\end{table}

\begin{table}[H]
    \centering
    \footnotesize
    \begin{tabular}{p{5cm}p{1cm}p{0.75cm}}
        \toprule
        \emph{Topic} & \emph{Model} & \emph{Variance} \\
        \midrule
        Suicide & QWEN & 0.099 \\
        Terrorism as a political, ideological or religious tactic & QWEN & 0.030 \\
        Euthanasia & QWEN & 0.031 \\
        \bottomrule
    \end{tabular}
    \vspace{2pt}
    \caption{\scriptsize Top 3 most controversial WVS topics according to QWEN}
    \label{tab:controversial_topicsWVSqwen}
\end{table}

\begin{table}[H]
    \centering
    \scriptsize
    \begin{tabular}{p{5.25cm}p{0.75cm}p{0.75cm}}
        \toprule
        \emph{Topic} & \emph{Model} & \emph{Variance} \\
        \midrule
        Suicide & BLOOM & 0.025 \\
        Terrorism as a political, ideological or religious tactic & BLOOM & 0.009 \\
        Euthanasia & BLOOM & 0.006 \\
        \bottomrule
    \end{tabular}
    \vspace{2pt}
    \caption{\scriptsize Top 3 most controversial WVS topics according to BLOOM}
    \label{tab:controversial_topicsWVSbloom}
\end{table}

\vspace{5pt}
\subsection{\small Most agreed on WVS topics according to models}
\vspace{-3pt}
\begin{table}[H]
    \centering
    \scriptsize
    \begin{tabular}{p{4cm}p{2cm}l}
        \toprule
        \emph{Topic} & \emph{Model} & \emph{Variance} \\
        \midrule
        Death penalty & GPT-2 Medium & 0.004 \\
        Accepting a bribe in the course of duty & GPT-2 Medium & 0.005 \\
        Parents beating children & GPT-2 Medium & 0.006 \\
        \bottomrule
    \end{tabular}
    \vspace{2pt}
    \caption{\scriptsize Top 3 most agreed on WVS topics according to GPT-2 Medium}
    \label{tab:agreed_topicsWVSgpt2-m}
\end{table}

\begin{table}[H]
    \centering
    \scriptsize
    \begin{tabular}{p{3.75cm}p{2cm}l}
        \toprule
        \emph{Topic} & \emph{Model} & \emph{Variance} \\
        \midrule
        Claiming government benefits to which you are entitled & GPT-2 Large & 0.002 \\
        Stealing property & GPT-2 Large & 0.004 \\
        Parents beating children & GPT-2 Large & 0.005 \\
        \bottomrule
    \end{tabular}
    \vspace{2pt}
    \caption{\scriptsize Top 3 most agreed on WVS topics according to GPT-2 Large}
    \label{tab:agreed_topicsWVSgpt2-l}
\end{table}

\begin{table}[H]
    \centering
    \scriptsize
    \begin{tabular}{p{3.75cm}p{2cm}l}
        \toprule
        \emph{Topic} & \emph{Model} & \emph{Variance} \\
        \midrule
        Claiming government benefits to which you are entitled & OPT-125 & 0.002 \\
        Someone accepting a bribe in the course of duty & OPT-125 & 0.003 \\
        For a man to beat his wife & OPT-125 & 0.004 \\
        \bottomrule
    \end{tabular}
    \vspace{2pt}
    \caption{\scriptsize Top 3 most agreed on WVS topics according to OPT-125}
    \label{tab:agreed_topicsWVSopt125}
\end{table}

\begin{table}[H]
    \centering
    \scriptsize
    \begin{tabular}{p{3.75cm}p{2cm}l}
        \toprule
        \emph{Topic} & \emph{Model} & \emph{Variance} \\
        \midrule
        Cheating on taxes & QWEN & 0.006 \\
        Homosexuality & QWEN & 0.007 \\
        Having casual sex & QWEN & 0.009 \\
        \bottomrule
    \end{tabular}
    \vspace{2pt}
    \caption{\scriptsize Top 3 most agreed on WVS topics according to QWEN}
    \label{tab:agreed_topicsWVSqwen}
\end{table}

\begin{table}[H]
    \centering
    \scriptsize
    \begin{tabular}{p{3.75cm}p{2cm}l}
        \toprule
        \emph{Topic} & \emph{Model} & \emph{Variance} \\
        \midrule
        Someone accepting a bribe in the course of duty & BLOOM & 0.001 \\
        Sex before marriage & BLOOM & 0.001 \\
        Avoiding a fare on public transport & BLOOM & 0.001 \\
        \bottomrule
    \end{tabular}
    \vspace{2pt}
    \caption{\scriptsize Top 3 most agreed on WVS topics according to BLOOM}
    \label{tab:agreed_topicsWVSbloom}
\end{table}

\subsection{\small Most controversial PEW topics according to models}

\begin{table}[H]
    \centering
    \scriptsize
    \begin{tabular}{p{3.75cm}p{2cm}l}
        \toprule
        \emph{Topic} & \emph{Model} & \emph{Variance} \\
        \midrule
        Getting a divorce & GPT-2 Medium & 0.043 \\
        Gambling & GPT-2 Medium & 0.039 \\
        Sex between unmarried adults & GPT-2 Medium & 0.023 \\
        \bottomrule
    \end{tabular}
    \vspace{2pt}
    \caption{\scriptsize Top 3 most controversial PEW topics according to GPT-2 Medium}
    \label{tab:controversial_topicsPEWgpt2_medium}
\end{table}

\begin{table}[H]
    \centering
    \scriptsize
    \begin{tabular}{p{3.75cm}p{2cm}l}
        \toprule
        \emph{Topic} & \emph{Model} & \emph{Variance} \\
        \midrule
        Sex between unmarried adults & GPT-2 Large & 0.059 \\
        Getting a divorce & GPT-2 Large & 0.043 \\
        Gambling & GPT-2 Large & 0.038 \\
        \bottomrule
    \end{tabular}
    \vspace{2pt}
    \caption{\scriptsize Top 3 most controversial PEW topics according to GPT-2 Large}
    \label{tab:controversial_topicsPEWgpt2_large}
\end{table}

\begin{table}[H]
    \centering
    \scriptsize
    \begin{tabular}{p{3.75cm}p{2cm}l}
        \toprule
        \emph{Topic} & \emph{Model} & \emph{Variance} \\
        \midrule
        Using contraceptives & OPT-125 & 0.041 \\
        Getting a divorce & OPT-125 & 0.038 \\
        Gambling & OPT-125 & 0.032 \\
        \bottomrule
    \end{tabular}
    \vspace{2pt}
    \caption{\scriptsize Top 3 most controversial PEW topics according to OPT-125}
    \label{tab:controversial_topicsPEWopt125}
\end{table}

\begin{table}[H]
    \centering
    \scriptsize
    \begin{tabular}{p{3.75cm}p{2cm}l}
        \toprule
        \emph{Topic} & \emph{Model} & \emph{Variance} \\
        \midrule
        Drinking alcohol & QWEN & 0.044 \\
        Getting a divorce & QWEN & 0.043 \\
        Gambling & QWEN & 0.027 \\
        \bottomrule
    \end{tabular}t
    \vspace{2pt}
    \caption{\scriptsize Top 3 most controversial PEW topics according to QWEN}
    \label{tab:controversial_topicsPEWqwen}
\end{table}

\begin{table}[H]
    \centering
    \scriptsize
    \begin{tabular}{p{3.75cm}p{2cm}l}
        \toprule
        \emph{Topic} & \emph{Model} & \emph{Variance} \\
        \midrule
        Sex between unmarried adults & BLOOM & 0.020 \\
        Using contraceptives & BLOOM & 0.006 \\
        Drinking alcohol & BLOOM & 0.006 \\
        \bottomrule
    \end{tabular}
    \vspace{2pt}
    \caption{\scriptsize Top 3 most controversial PEW topics according to BLOOM}
    \label{tab:controversial_topicsPEWbloom}
\end{table}

\subsection{\small Most agreed on PEW topics according to models}

\begin{table}[H]
    \centering
    \scriptsize
    \begin{tabular}{p{3.75cm}p{2cm}l}
        \toprule
        \emph{Topic} & \emph{Model} & \emph{Variance} \\
        \midrule
        Having an abortion & GPT-2 Medium & 0.013 \\
        Homosexuality & GPT-2 Medium & 0.016 \\
        Using contraceptives & GPT-2 Medium & 0.016 \\
        \hline
    \end{tabular}
    \vspace{2pt}
    \caption{\scriptsize Top 3 most agreed on PEW topics according to GPT-2 Medium}
    \label{tab:agreed_topicsPEWgpt2_medium}
\end{table}

\begin{table}[H]
    \centering
    \scriptsize
    \begin{tabular}{p{3.75cm}p{2cm}l}
        \toprule
        \emph{Topic} & \emph{Model} & \emph{Variance} \\
        \midrule
        Having an abortion & GPT-2 Large & 0.015 \\
        Using contraceptives & GPT-2 Large & 0.023 \\
        Drinking alcohol & GPT-2 Large & 0.023 \\
        \bottomrule
    \end{tabular}
    \vspace{2pt}
    \caption{\scriptsize Top 5 most agreed on PEW topics according to GPT-2 Large}
    \label{tab:agreed_topicsPEWgpt2_large}
\end{table}

\begin{table}[H]
    \centering
    \scriptsize
    \begin{tabular}{p{3.75cm}p{2cm}l}
        \toprule
        \emph{Topic} & \emph{Model} & \emph{Variance} \\
        \midrule
        Married people having an affair & OPT-125 & 0.011 \\
        Homosexuality & OPT-125 & 0.019 \\
        Sex between unmarried adults & OPT-125 & 0.020 \\
        \bottomrule
    \end{tabular}
    \vspace{2pt}
    \caption{\scriptsize Top 3 most agreed on PEW topics according to OPT-125}
    \label{tab:agreed_topicsPEWopt125}
\end{table}

\begin{table}[H]
    \centering
    \scriptsize
    \begin{tabular}{p{3.75cm}p{2cm}l}
        \toprule
        \emph{Topic} & \emph{Model} & \emph{Variance} \\
        \midrule
        Having an abortion & QWEN & 0.004 \\
        Married people having an affair & QWEN & 0.006 \\
        Using contraceptives & QWEN & 0.006 \\
        \bottomrule
    \end{tabular}
    \vspace{2pt}
    \caption{\scriptsize Top 3 most agreed on PEW topics according to QWEN}
    \label{tab:agreed_topicsPEWqwen}
\end{table}

\begin{table}[H]
    \centering
    \scriptsize
    \begin{tabular}{p{3.75cm}p{2cm}l}
        \toprule
        \emph{Topic} & \emph{Model} & \emph{Variance} \\
        \midrule
        Getting a divorce & BLOOM & 0.002 \\
        Homosexuality & BLOOM & 0.003 \\
        Gambling & BLOOM & 0.004 \\
        \bottomrule
    \end{tabular}
    \vspace{2pt}
    \caption{\scriptsize Top 3 most agreed on PEW topics according to BLOOM}
    \label{tab:agreed_topicsPEWbloom}
\end{table}

\end{document}